**ORIGINAL PAPER**

# Hybrid thermal modeling of additive manufacturing processes using physics-informed neural networks for temperature prediction and parameter identification

Shuheng Liao[1] · Tianju Xue[1] · Jihoon Jeong[1] · Samantha Webster[1] · Kornel Ehmann[1] · Jian Cao[1]

**Abstract**

Understanding the thermal behavior of additive manufacturing (AM) processes is crucial for enhancing the quality control and enabling customized process design. Most purely physics-based computational models suffer from intensive computational costs and the need of calibrating unknown parameters, thus not suitable for online control and iterative design application. Data-driven models taking advantage of the latest developed computational tools can serve as a more efficient surrogate, but they are usually trained over a large amount of simulation data and often fail to effectively use small but high-quality experimental data. In this work, we developed a hybrid physics-based data-driven thermal modeling approach of AM processes using physics-informed neural networks. Specifically, partially observed temperature data measured from an infrared camera is combined with the physics laws to predict full-field temperature history and to discover unknown material and process parameters. In the numerical and experimental examples, the effectiveness of adding auxiliary training data and using the pretrained model on training efficiency and prediction accuracy, as well as the ability to identify unknown parameters with partially observed data, are demonstrated. The results show that the hybrid thermal model can effectively identify unknown parameters and capture the full-field temperature accurately, and thus it has the potential to be used in iterative process design and real-time process control of AM.

## 1 Introduction

Metal additive manufacturing (AM) processes have been increasingly utilized in various industries like aerospace, automotive, and biomedical industries because of their high flexibility in fabricating complex geometries with various materials. Directed energy deposition (DED) and Powder bed fusion (PBF), are the two major types of metal additive manufacturing processes. In both DED and PBF, a high energy laser moves along a pre-defined toolpath and melts the material to build the part in a layer-by-layer manner. Understanding the highly transient thermal behavior of AM processes can help to maintain the process stability and part quality because it is strongly related to the process defects such as lack of fusion, porosity, and cracking. In addition, thermal history is also a key link in the process–structure–property chain of the material in AM processes [1]. Therefore, developing efficient and reliable thermal models of AM processes is crucial for enhancing the process quality control as well as for process design.

Physics-based computational models using finite element method (FEM) and computational fluid dynamics (CFD) have been widely studied to predict thermal behavior in AM processes. Thermal-fluid models [2,3], considering the heat transfer in the entire part and fluid flow within the melt pool by solving the Navier–Stokes equations including energy conservation, are considered as high-fidelity computational models for predicting the thermal behavior. Gan et al. [4] improved the thermal-fluid model by considering heat loss due to vaporization. It was shown through the NIST AM benchmark exercise [5,6] that the model with vaporization reduced the average difference between the numerical and experimental cooling rates during solidification from 28 to 12% compared to that without fluid flow and vaporization. However, the difference in the cooling rate after solidifica-tion does not vary too much. In most cases, the CFD models are computationally expensive and are usually performed at mesoscale. FEM models [7,8], considering only the solid heat transfer in AM processes, are computationally less expensive than CFD models and thus can be applied to the simulation of macroscopic distortion [9] and residual stress [10,11], as well as offline process design and planning [12,13]. However, these FEM models are still too time-consuming for time-sensitive applications like online process control. In addition, using conventional physics-based models requires accurate input of material and process parameters, which is usually costly to obtain from experiments, to avoid significant discrepancies between the model and actual experimental conditions.

✉ Jian Cao
  jcao@northwestern.edu

[1]  Department of Mechanical Engineering, Northwestern University, Evanston 60208, IL, USA

In the past few years, data-driven prediction models in advanced manufacturing have gained increasing attention because of their potential to derive end-to-end models directly from big data and improve computational efficiency with the latest computational tools based on graphic processing units (GPUs). In our early work [14], a data-driven model based on recurrent neural networks (RNNs) was developed to predict the thermal history from the time-series toolpath features, geometric features and laser features, and the results showed that the model reaches less than $3\mathrm{e}-5$ normalized mean squared error (MSE) after the training of 100 epochs with the dataset generated from FEM simulations. More recently, Roy et al. [15] developed a machine learning model of thermal history in AM. In their model, the G-code is directly translated into a set of input features, e.g., distance from heat source and cooling surface, for predicting the local thermal history. The model was trained with FEM simulation data and the prediction error was less than 5% in almost real time. Ren et al. [16] developed a data-driven model predicting the thermal field evolution of a single layer built from different toolpath strategies in AM, where 100 different one-layer deposition cases with 6 different toolpath strategies were simulated using FEM for generating the dataset and the model achieved a prediction accuracy of 95% after the training. Zhou et al. [17] extended the idea and developed the thermal field prediction model for arbitrary multi-layer geometries by proposing a novel method for the discretization of the deposition process, and achieved an accuracy exceeding 94% in the validation dataset. In our recent work [18], a geometry-agnostic thermal model based on graph neural networks (GNNs) was proposed where the FEM simulation data was converted into a graph representation to train the GNN model. With the dataset generated from 50 different geometries, the trained model can predict the thermal history for unseen geometries with a root-mean-square error (RMSE) less than $1\mathrm{e}-2$. Although data-driven models have been demonstrated to be computationally efficient, these models require a large amount of labelled training data, which is often impractical to be obtained from the exper-iment. Most of the studies listed above, and other previous studies [19,20] used FEM simulation to generate the training dataset, in which case although the prediction accuracy is good compared with the simulation data, the models are still not directly connected with the actual experimental conditions and the accuracy strongly relies on the fidelity of the simulation models.

In this work, we developed a hybrid (both physics-based and data-driven) framework that allows to arbitrarily fuse "big data" (simulations) and/or "small data" (experiments) into a physics-informed model, hence providing a customized, flexible platform for efficient thermal analysis in AM. Our approach is based on physics-informed neural networks (PINNs), originally proposed by Raissi et al. [21] as a surrogate model to solve partial differential equations (PDEs). PINNs have been widely used to solve forward problems, i.e., solving PDEs, and inverse problems, i.e., identifying parameters, in engineering applications like fluid mechanics [22,23], heat transfer [24,25], and solid mechanics [26,27]. PINNs provide a way to define the loss function from the governing equations. By combining the PINN model with the labelled data, a hybrid model can be developed which takes advantage of physics laws to eliminate the need for large training datasets, and in return, allowing the discovery of unknown physics from the data. In a recent work by Zhu et al. [28], PINNs were applied to predict the thermal field in AM, where the physical parameters were assumed to be known, and the FEM simulation data was used to train the PINN model. Here, we developed a hybrid framework for AM processes using PINNs for the first time, where partially observed temperature data measured from an infrared (IR) camera is combined with the physics laws to predict the full-field temperature history. We also demonstrate that it can be used to discover unknown material and process parameters.

The rest of the paper is organized as follows. In Sect. 2, the governing equation of the heat transfer problem and the basic formation of PINNs are introduced. In Sect. 3, a numerical example of bare plate scanning is performed where the effect of auxiliary data and pretraining in solving forward problems are studied. The ability of the hybrid model to identify unknown parameters from partially observed data is demonstrated in Sect. 4. In Sect. 5, an experimental case of a DED thin wall part, where PINN is used to predict the full-field temperature data is presented. Conclusions and future work are given in Sect. 6.

## 2 Methods

Figure 1 illustrates the developed hybrid PINN-based framework for thermal modeling of AM processes. The neural network takes the spatial-temporal coordinates as the input and predicts the temperature at the corresponding point. The loss function for training the neural network is defined as the combination of the residuals of the PDEs, the boundary conditions (BCs) and the initial conditions (ICs), as well as an extra data-based loss term where the experimental temperature data is used as the ground truth. In the following subsections, the governing equations of the heat transfer problem in AM and the basic formation of PINNs are introduced.

## 2.1 Governing equations

In this work, to demonstrate the hybrid framework, we model only the heat conduction in AM, while fluid flow and vaporization heat loss are ignored. It is noted that adding the implementation of these additional physics in the framework will be relatively easy since PINN is a meshless optimization-based approach compared to the conventional finite element or finite difference method. The governing equation of the transient heat conduction in AM can be written as:

$$\rho C_p \frac{\partial T}{\partial t} + \nabla \cdot \mathbf{q} = 0, \tag{1}$$

where $\rho$ is the density of the material, $C_p$ is the heat capacity, $T$ is temperature, $t$ is time and $\mathbf{q}$ is the heat flux, which is given by the Fourier's Law:

$$\mathbf{q} = -k \nabla T, \tag{2}$$

where $k$ is the thermal conductivity of the material. The latent heats of fusion and vaporization of the material are not considered in the current work. The heat flux boundary condition in AM processes can be described as:

$$\mathbf{q} \cdot \mathbf{n} = q_{\text{laser}} + q_{\text{conv}} + q_{\text{rad}}, \tag{3}$$

where $q_{\text{laser}}$, $q_{\text{conv}}$, and $q_{\text{rad}}$ are the heat flux due to the laser heat source, convection and radiation, respectively. The laser heat flux is modeled using a Gaussian surface heat flux model:

$$q_{\text{laser}} = -\frac{2\eta P}{\pi r_{\text{beam}}^2} \exp\left(\frac{-2d^2}{r_{\text{beam}}^2}\right), \tag{4}$$

where $\eta$ is the laser absorptivity, $P$ is the laser power, $r_{\text{beam}}$ is the laser beam radius, and $d$ is the distance from the material point to the laser center. The convective and radiative heat flux can be calculated by:

$$q_{\text{conv}} = h(T - T_0), \tag{5}$$

$$q_{\text{rad}} = \sigma \varepsilon (T^4 - T_0^4), \tag{6}$$

where $h$ is the convection heat transfer coefficient, $\sigma$ is the Stefan-Boltzmann constant, $\varepsilon$ is the emissivity of the material and $T_0$ is the ambient temperature. The laser heat flux is assumed to be applied on only the top surface while the convective and radiative heat flux are applied on all surfaces except for the bottom surface of the substrate, where a Dirichlet boundary condition is applied:

$$T \mid_{z=0} = T_0. \tag{7}$$

## 2.2 Physics-informed neural networks

PINNs are neural networks with physics information encoded. The basic idea of PINN is to approximate the solution of PDEs using a neural network, which is trained by minimizing a loss function defined based on the residuals of the PDEs, BCs, and ICs. Considering a PDE defined in the spatial-temporal domain of the general form:

$$u_t + \mathcal{N}[u] = 0, \ \mathbf{x} \in \Omega, t \in [0, T], \tag{8}$$

where $u(\mathbf{x}, t)$ is the solution of the PDE, $\mathcal{N}[\cdot]$ is a general differential operator; $\mathbf{x}$ and $t$ are the spatial and temporal coordinates, $\Omega$ is the computational domain which is a subset of $\mathbb{R}^3$. The BCs and ICs of the PDE can be written as:

$$\mathcal{B}(u, \mathbf{x}, t) = 0, \ \mathbf{x} \in \partial \Omega, \tag{9}$$

$$u(\mathbf{x}, 0) - I(\mathbf{x}) = 0, \ \mathbf{x} \in \Omega, \tag{10}$$

where $\partial \Omega$ is the boundary of the computation domain. In the framework of PINN, $u$ is approximated using a fully connected neural network, of which the input is $\mathbf{x}$ and $t$, and the output is the approximated solution $\hat{u}(\mathbf{x}, t)$. The loss function of PINN is defined as:

$$\mathcal{L} = w_b \mathcal{L}_b + w_i \mathcal{L}_i + w_r \mathcal{L}_r, \tag{11}$$

where $\mathcal{L}_b$, $\mathcal{L}_i$, and $\mathcal{L}_r$ are the losses from BCs, ICs and the PDE residuals, respectively, and $w_b$, $w_i$ and $w_r$ are the weights of each term. When training a PINN, sampling points are first selected at the boundary, at the initial state, and in the temporal-spatial domain. At each iteration, each loss term can be calculated at its respective sampling points:

$$\mathcal{L}_b = \frac{1}{N_b} \sum_{k=1}^{N_b} \left| \mathcal{B}(\hat{u}(\mathbf{x}_b^k, t_b^k), \mathbf{x}_b^k, t_b^k) \right|^2, \tag{12}$$

$$\mathcal{L}_i = \frac{1}{N_i} \sum_{k=1}^{N_i} \left| \hat{u}(\mathbf{x}_i^k, 0) - I(\mathbf{x}_i^k) \right|^2, \tag{13}$$

$$\mathcal{L}_r = \frac{1}{N_r} \sum_{k=1}^{N_r} \left| \hat{u}(\mathbf{x}_r^k, t_r^k) - \mathcal{N}[\hat{u}] \right|^2, \tag{14}$$

where $N_b$, $N_i$, and $N_r$ are the number of the sampling points for each loss term. PINNs use automatic differentiation (AD), a method commonly used in neural network applications [29] and integrated with most of the deep learning packages like TensorFlow [30] and PyTorch [31], to automatically evaluate the spatial and temporal derivatives of the PDE. This is in contrast with conventional methods such as the finite difference method, where derivatives are only approximated with numerical difference.

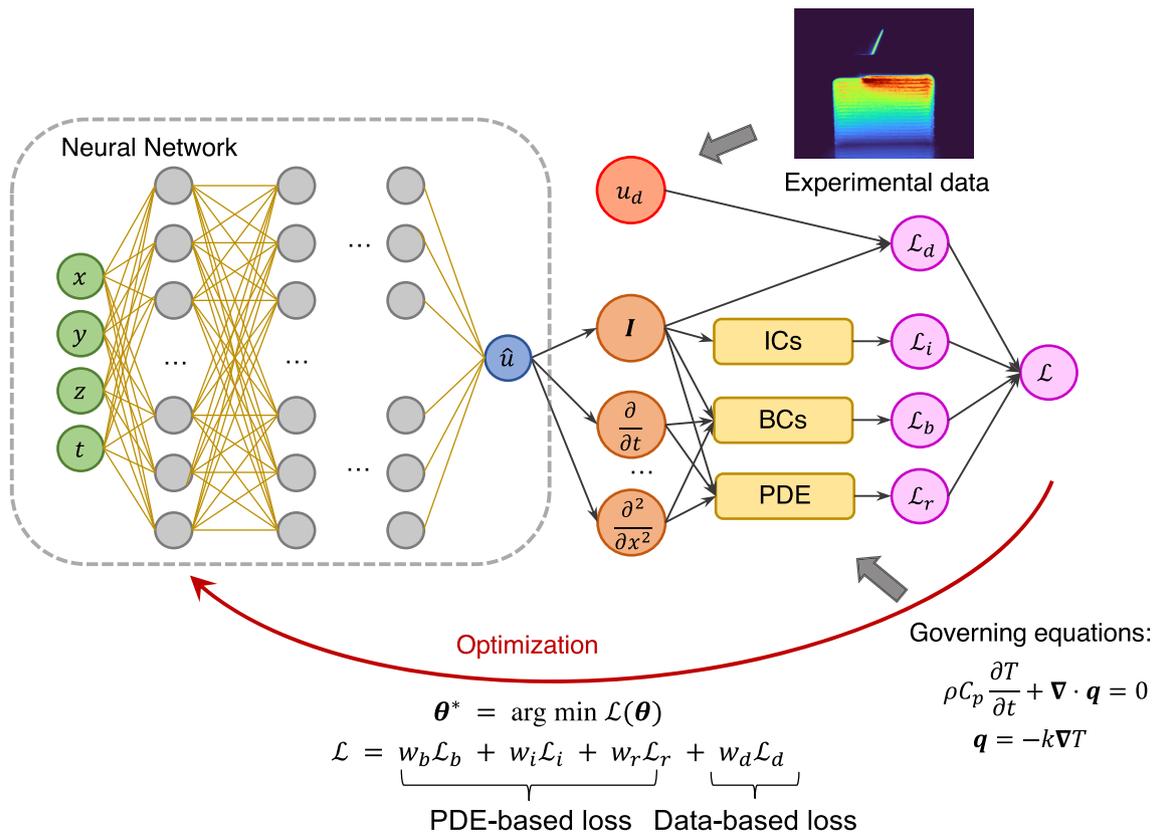

**Fig. 1** Hybrid thermal modeling framework for AM based on PINNs

Based on Eqs. (11) to (14), the PINN model can be trained to solve PDEs iteratively like an optimization problem, which is known as the forward problem. In addition, when extra labelled data is available, a data-based loss term can be added to the total loss of the PINN. The data-based loss is like the loss in the conventional supervised learning scheme:

$$\mathcal{L}_d = \frac{1}{N_d} \sum_{k=1}^{N_d} \left| \hat{u}(\boldsymbol{x}_i^d, t_i^d) - u(\boldsymbol{x}_i^d, t_i^d) \right|^2, \tag{15}$$

and then

$$\mathcal{L} = w_b \mathcal{L}_b + w_i \mathcal{L}_i + w_r \mathcal{L}_r + w_d \mathcal{L}_d. \tag{16}$$

Because of the nature of the PINN, both the physics laws and data are "soft" constraints to the model, meaning that the loss usually decreases during the training but will not be exactly zero. Since the constraints are not strictly satisfied, the labelled data can be arbitrarily fused with the physics laws without leading to an over-constrained problem. There is no specific requirement for the size and fidelity of the labelled data, therefore, such data could be of any simulation results, or measured data from the experiments.

The idea of adding a data-based loss has the following three advantages: (1) the labelled data can serve as auxiliary data to guide the training of PINNs and thus can accelerate the training to solve *forward problems*; (2) it can also be used to solve the *inverse problems*, i.e., identify unknown parameters or discovery physics laws; and (3) it enables arbitrary fusion of the experimental data into the physics laws to create a *hybrid physics-informed model*. In the following sections, numerical and experimental examples are providing to demonstrate the three cases, respectively.

## 3 Solving the forward problem

To demonstrate the effectiveness of the PINN model in solving forward problems of AM processes, we first use FEM simulation results as the exact solution to benchmark the performance of the prediction results using PINN. Here, a numerical example of a bare plate scanning case is presented.

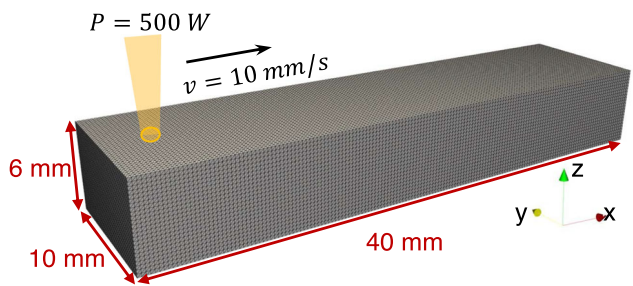

**Fig. 2** Schematic of the numerical example: the simulation setup

**Table 1** Simulation parameters of the numerical example

| Parameter | Value | Unit |
|-----------|-------|------|
| Density, $\rho$ | 8 | g/cm$^3$ |
| Heat capacity, $C_p$ | 0.5 | J/(g K) |
| Heat conductivity, $k$ | 10 | W/(m K) |
| Convection coefficient, $h$ | 20 | W/(m$^2$ K) |
| Emissivity, $\varepsilon$ | 0.3 | – |
| Laser absorptivity, $\eta$ | 0.4 | – |
| Ambient temperature, $T_0$ | 298 | K |

The dimension of the bare plate was assumed to be 40 mm * 10 mm * 6 mm. A 500 W laser of 1.5 mm beam radius was used to scan the bare plate with a speed of 10 mm/s, as shown in Fig. 2. The numerical example was solved using the FEM implemented in FEniCS [32]. A tetrahedron mesh of 0.25 mm element size was used in the simulation. The mesh consists of a total of 921,600 elements and 165,025 nodes. The Crank-Nicolson method, which is an implicit time integration scheme combining the forward and backward Euler methods [33], was used for time integration. The time step size was set as 5 ms. A total of 3 seconds of the process was simulated. The simulation parameters of this numerical example are shown in Table 1. The temperature value at each node with a frequency of 10 Hz was extracted from the simulation (4,950,750 data points in total). The simulation took about 4 h on an Intel Xeon Silver 4110 CPU using a single core.

## 3.1 PINN trained without auxiliary data

As discussed in the previous section, PINN can be used as a surrogate model to solve the PDE using the PDE-based loss function defined by Eqs. (11) to (14). The PINN used in this example consists of 3 hidden layers with 64 neurons in each layer. The input of the PINN model is the 4-dimensional spatial-temporal coordinates, which are scaled to [−1, 1], while the output is the temperature normalized to [0, 1]. The hyperbolic tangent function is used as the activation function for all the layers except for the output layer where the Soft-plus function is used to ensure a positive output, so that the predicted temperature can always be larger than the ambient temperature considering the physics of the process.

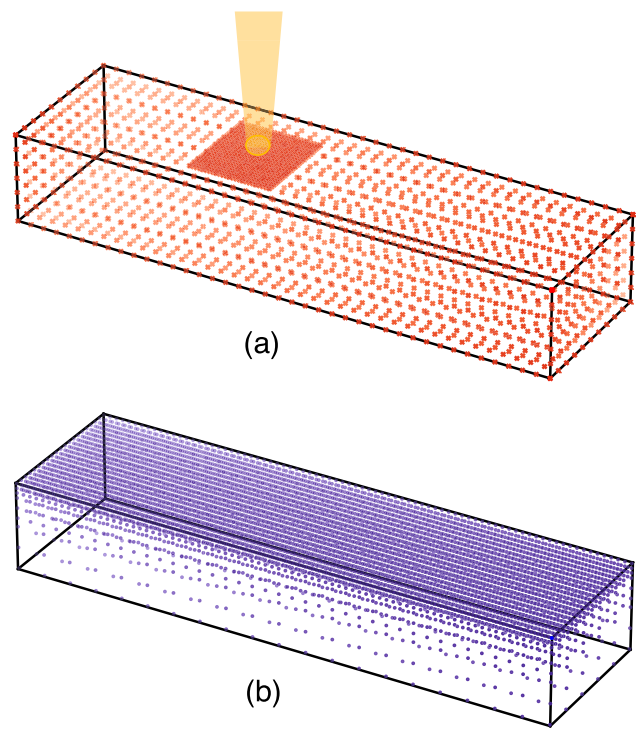

**Fig. 3** Sampling points at time $t$ **a** on the boundary; **b** in the domain

In order to form the PDE-based loss, sampling points need to be selected at the boundary, at the initial state and in the domain to calculate the residuals in Eqs. (12) to (14). In order to capture the highly transient boundary conditions in AM, the spatial-temporal domain was first discretized with a uniform time step of 0.05 s. At each time step, a uniform 2D grid of 1 mm interval was sampled in each surface. In particular, extra points with 0.25 mm interval were sampled in the 6 mm * 6 mm area near the laser center, as shown in Fig. 3a. For the residual points in the domain, it is suggested in [34] that putting more points near the region where the gradient is large can help the training of PINN. In AM processes, it is known that the temperature gradient is large in the top layers near the laser center. Therefore, in this example we used a uniform grid of 0.5 mm within the 40 mm * 10 mm * 1 mm top region, which is 4 times finer than that in the lower region.

The PINN model was implemented using PyTorch and was first trained without labelled data for 50,000 epochs using an Adam optimizer [35], which took about 2.5 h on an RTX A6000 GPU. A learning rate of 2e−4 was used in the training. The performance of the trained PINN model is compared with the benchmark FEM results. Figure 4 presents the comparisons between the temperature field at different time steps predicted by the PINN model and the FEM simulation from the top view and the cross-section view. It is observed that the

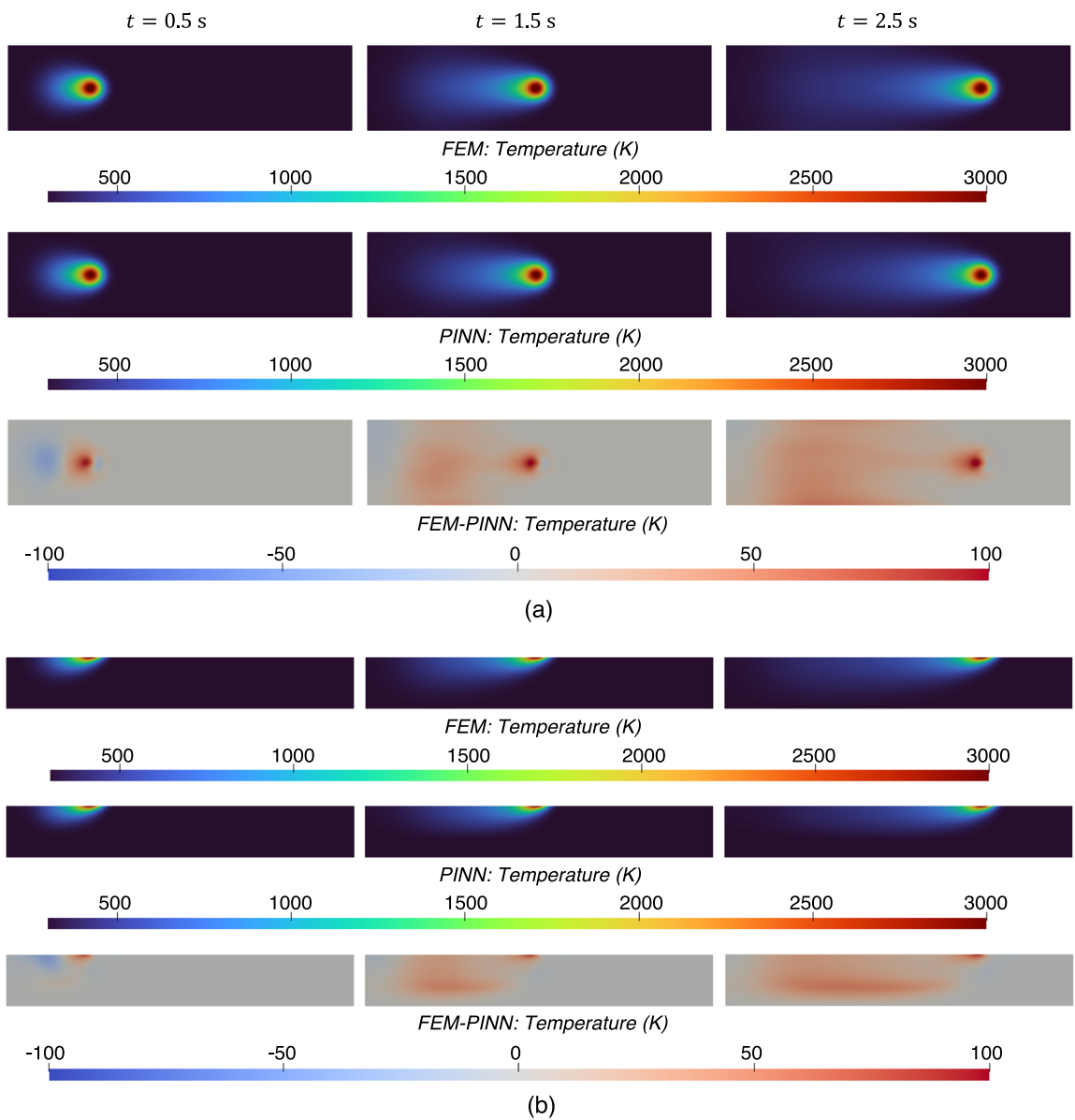

$t = 0.5$ s   $t = 1.5$ s   $t = 2.5$ s

**Fig. 4** Comparison of the temperature field from benchmark FEM simulation and the trained PINN model: **a** top view and **b** cross-section view

PINN predicted temperature field matches well with the FEM results, with the RMSE of 14.07 K. The evolution of each loss term during the training process is shown in Fig. 5a. At the beginning of the training, the PDE residual loss is small due to that Eq. (1) is relatively easy to be satisfied with a uniform temperature field, while the BCs defined in Eqs. (4) to (7) are not satisfied at this stage. As the training proceeds, the neural network gradually learns the initial conditions and boundary conditions of the system and tends to maintain the three loss terms at the same order.

## 3.2 Use of the auxiliary data for accelerating the training

In the next step, the effect of adding the auxiliary data on the rate of convergence of the training of PINN was studied. Here, we used the simulation data as the auxiliary data to accelerate the training of the PINN model. 100,000 labelled data points were randomly selected from the benchmark FEM results to form the data loss defined in Eq. (15). Two cases using the clean data and the noisy data with a Gaussian noise

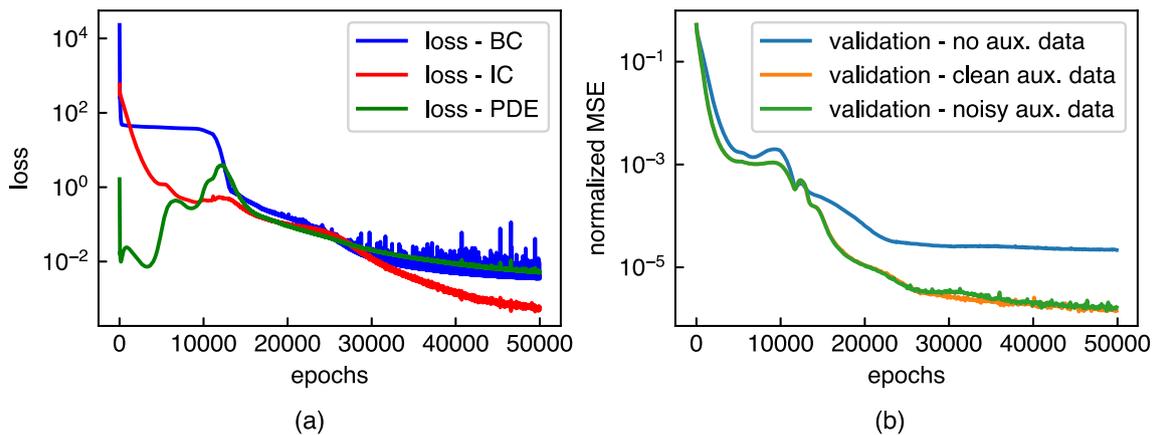

**Fig. 5** **a** Evolution of each loss term for the PINN trained without data, **b** comparison of the evolutions of the normalized MSE error of the PINN models trained with/without auxiliary data

of standard deviation of 100 K added were tested. The entire dataset from the benchmark FEM simulation was used for validation. The same network structure and training parameters as in the case without auxiliary data were used. Figure 5b is the comparison of the evolution of the normalized MSE during the PINN training without auxiliary data, with clean data, and with noisy data, respectively. It shows that the auxiliary data can accelerate the training of PINN, and eventually improve the prediction accuracy. Interestingly, the performance in the case using the noisy data is very close to that in the case using clean data, showing that PINN is robust and accepts relatively low-quality data to improve the training process. The RMSE compared with FEM results in the three cases are shown in Table 2. With auxiliary data, the PINN model reached the same accuracy (about 2.2e−5 normalized MSE) with about 1/3 of the number of epochs compared to that without auxiliary data.

### 3.3 Effects of the initial guess

Since PINN essentially solves a non-convex optimization problem, the success of the training process is strongly affected by the initial guess of the bias and weights of the neural network. If the initial guess is close to the optimal solution, then the neural network can be trained with many fewer epochs [36]. Therefore, compared with conventional numerical methods, one advantage of using PINN to solve PDEs is that the past results can be re-utilized as a pretrained model when the parameters in the PDEs are changed.

As an example, the PINN model trained without auxiliary data discussed in the previous section is used as the pretrained model to solve an AM problem with a similar setup (see Fig. 2), where the process parameter is changed from $P = 500$ W and $v = 10$ mm/s to $P = 400$ W and $v = 8$ mm/s. Figure 6a shows the evolutions of each loss term of the loss function during the 10,000 training epochs. One can see that in this case with the pretrained model the loss function converges faster than that in Fig. 5a. Figure 6b shows the comparison of the normalized MSE during the training of the PINN started from the pretrained model and a randomly initialized model. The FEM solution of the case with $P = 400$ W and $v = 8$ mm/s is used as the baseline to calculate the MSE here. The PINN started from the pretrained model took less than 1/5 of the number of epochs in the case of a randomly initialized model to reach the same accuracy, hence demonstrating the transfer learning capabilities of PINNs when solving the same problem with multiple different process or material parameters, e.g., in process design applications.

## 4 Solving the inverse problem

In AM processes, the transient thermal behavior is strongly related to not only the material properties, i.e., heat capacity, thermal conductivity, density, etc., but also the experimental conditions like laser absorptivity, convection coefficient, and ambient temperature, of which the related parameters usually need to be calibrated when the hardware setup or the environment is changed. In [21], it is demonstrated that PINN has the ability to identify the unknown values of the parameters in PDEs by optimizing the unknown parameters and the weights and bias of the neural network at the same time:

$$\boldsymbol{\theta}^*, \boldsymbol{\mu}^* = \arg\min \mathcal{L}(\boldsymbol{\theta}, \boldsymbol{\mu}), \tag{17}$$

where $\boldsymbol{\mu}$ is the vector of unknown parameters, and $\boldsymbol{\theta}$ is the vector of weights and bias of the neural network. Here, a synthetic case is presented where the developed PINN-based thermal model can be used to identify unknown material and process parameters from partially observed sparse temperature data.

| Table 2 Comparison of the performance of PINNs trained with and without auxiliary data | | No auxiliary data | Clean auxiliary data | Noisy auxiliary data |
|---|---|---|---|---|
| | RMSE (K) | 14.07 | 3.59 | 3.72 |

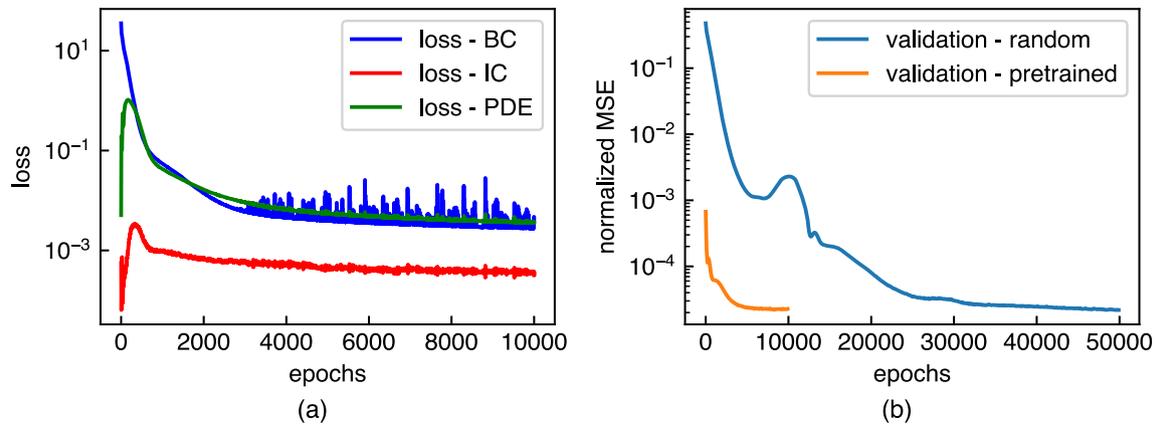

(a)  (b)

**Fig. 6** **a** evolution of each loss term in the case of starting from the pretrained model; **b** comparison of the evolutions of the normalized MSE error of the PINN models trained from the pretrained model and from a randomly initialized model

It is common to use a coaxial IR camera to monitor the melt pool temperature in PBF or DED processes to get the basic knowledge of the thermal behavior during the process and control the quality of the fabricated part. Using the partially observed temperature field data from a coaxial IR camera, the ability of PINN to identify the unknown parameters and to infer the full-field temperature data was tested. In this case, synthetic IR images were generated from the FEM simulation results described above. The temperature data was extracted in the 6 mm * 6 mm range around the laser center with a spatial resolution of 0.25 mm and a frequency of 10 Hz. Pixels with temperatures higher than 2000 K were removed to imitate the high temperature saturation of the IR camera in the actual experiment. Gaussian noise of a standard deviation of 100 K was added to the synthetic images. Figure 7 shows the generated synthetic IR images of the melt pool.

In the first example, laser absorptivity $\eta$ was assumed to be unknown, and the PINN model was trained from randomly initialized weights and bias for 50,000 epochs. Figure 8a is the evolution of each loss term while Fig. 8b is the history of the normalized MSE error compared to the benchmark FEM result and the value of $\eta$ during the training process. The final identified value of $\eta$ was 0.404, which is close to the ground truth value of 0.400. Also, the RMSE of the temperature in the full field is 11.45 K, which is comparable to the values in Table 2, showing that using PINN the full-field temperature can be well predicted from the partially observed sparse temperature data in AM processes with unknown process parameters.

In the second example, it was assumed that the material properties, i.e., $C_p$ and $k$ are unknown. The evolution of each loss term and $C_p$ and $k$ values during the training are shown in Fig. 9. Since the initial values of $C_p$ and $k$ were set to to 0 at the beginning of the training and then kept increasing. After about 10,000 epochs the four loss terms reached about the same magnitude and then gradually decreased at the same time. The final identified $C_p$ value is 0.512 J/(g K) and $k$ value is 9.51 W/(m K), close to the ground truth values $C_p = 0.500$ J/(g K) and $k = 10.0$ W/(m K).

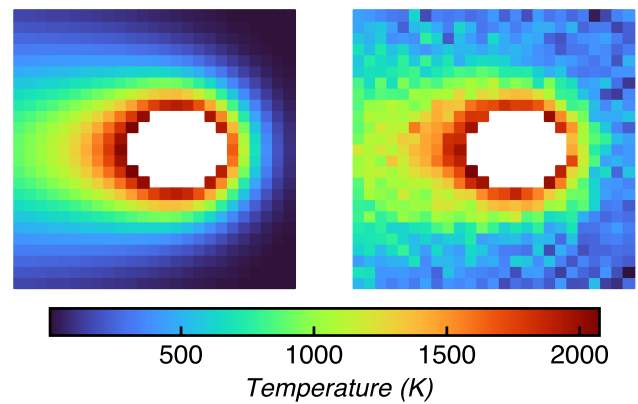

**Fig. 7** Generated synthetic melt pool images no noise (left) and with noise (right)

# 5 Hybrid model for full-field temperature prediction

In AM processes, it is common to use IR cameras [37–39] to monitor the melt pool temperature and geometry, where the temperature field can be partially observed. With the PINN-based hybrid thermal model, as Fig. 1 illustrates, the temperature history in the entire domain can be inferred from the partially observed temperature data and the governing equation of the process. In this section, an experimental case is presented where the developed PINN-based thermal model of AM processes was applied to an actual DED experiment to predict the full-field temperature. Specifically, we assume the temperature field at the heat-affected zone can be measured from the experiment so that this hybrid model can eliminate the need of calibrating the heat source model, i.e., beam profile and absorption rate, in conventional FEM models.

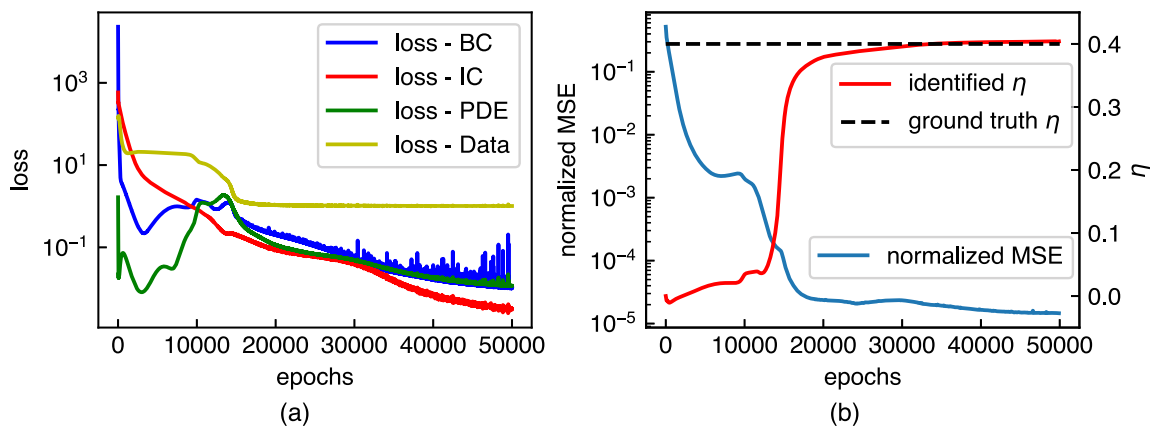

**Fig. 8** **a** Evolution of each loss term, **b** history of the normalized MSE error compared to benchmark FEM result and the value of $\eta$ during the training process

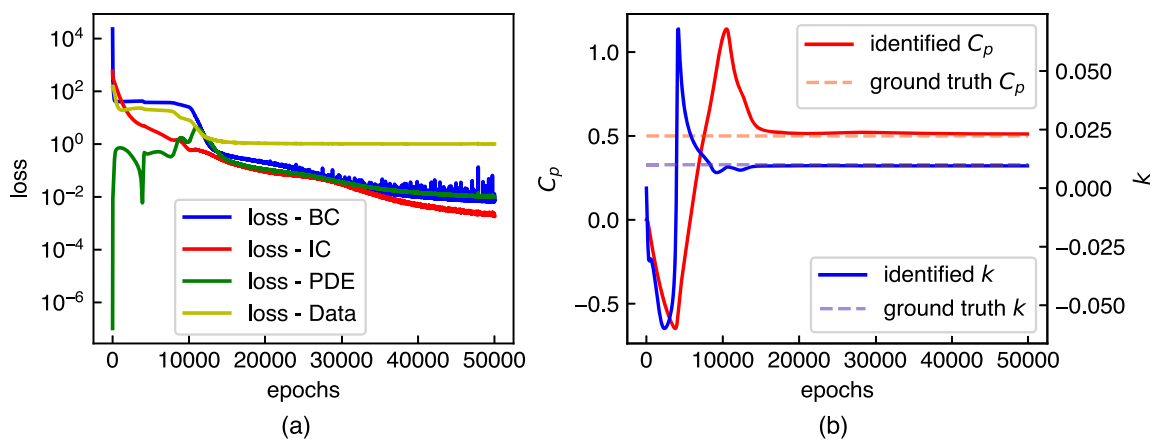

**Fig. 9** **a** Evolution of each loss term, **b** history of $C_p$ and $k$ values during the training process

## 5.1 Experimental setup

The experiment was performed on a customized DED system, i.e., the Additive Rapid Prototyping Instrument (ARPI), developed at Northwestern University. The primary processing laser of the system is an IPG Photonics fiber laser with a 1 kW (1070 nm) maximum laser power and a 1.12 mm beam radius (measured $1/e^2$ value). The experimental setup is shown in Fig. 10. During the experiment, a FLIR A655sc IR camera with a resolution of 480 * 640 pixels was used to capture the temperature field from the side view at a rate of 50 Hz.

In the experiment, a 40-layer thin wall part was built. Figure 11 depicts the size of the built thin wall and the substrate. The thin wall part was built using a bi-directional toolpath with a 7 mm/s scanning speed and a 500 W laser power. A 0.5 second dwell time was used between the deposition of each layer. The substrate material is AISI 1018 steel while the part material is the nickel-based alloy IN718.

The FLIR IR camera was used in the experiment to measure the temperature of the side surface of the wall (about 0.2 mm spatial resolution) during the experiment. The workflow of processing the IR data is shown in Fig. 12. A constant emissivity value was assumed to calculate the temperature values. The raw data was first cropped and then downsampled to 60*40 (0.75 mm spatial resolution) by averaging the surrounding pixels to reduce the noise of the measurement.

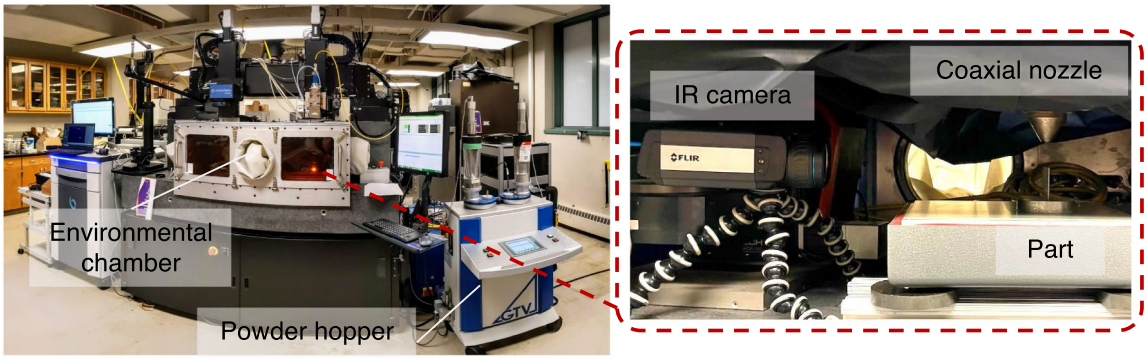

**Fig. 10** Experimental setup: overview of the ARPI system (left) and the temperature measurement setup (right)

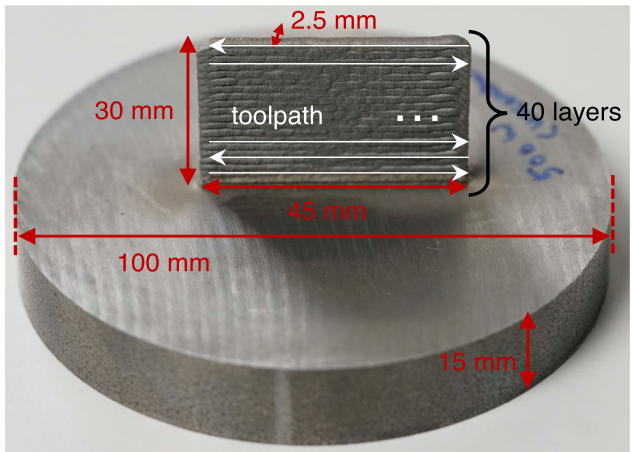

**Fig. 11** The built wall part from the experiment

## 5.2 Full field temperature prediction

In this example, the developed PINN-based hybrid model was used to infer the full temperature field from the partially observed temperature data. One basic assumption we made in this example is that the temperature is uniform along the through-the-thickness direction and the solid heat transfer through this direction can be ignored. With this assumption,

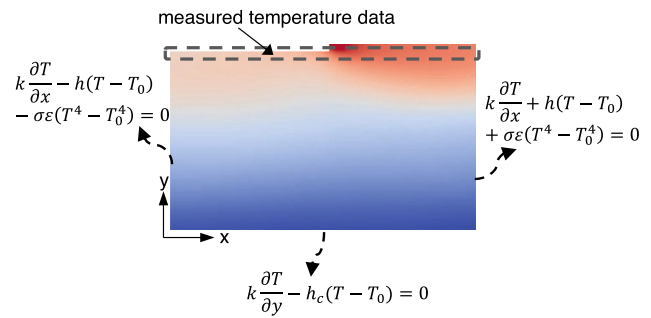

**Fig. 13** Illustration of the simplified 2D problem of the thin wall part with boundary conditions highlighted by equations

the heat transfer in the thin wall case can be simplified to a 2D problem, as Fig. 13 illustrates.

The convective and radiative heat flux on the two surfaces of the wall parallel to the $xy$ plane is treated as a heat source term. Equation 1 can be then written as:

$$\rho C_p \frac{\partial T}{\partial t} - k\nabla^2 T + \frac{2h}{w}(T - T_0) + \frac{2\sigma\varepsilon}{w}(T^4 - T_0^4) = 0,$$
(18)

where $w$ is the thickness of the wall, i.e., 2.5 mm in this case. The boundary condition in Eq. (3) still applies to the left and right boundaries of the 2D wall. For the bottom of the wall, since the temperature of the substrate is assumed to be always equal to the ambient temperature, a flux boundary condition is assumed:

$$k\frac{\partial T}{\partial y} - h_c(T - T_0) = 0, \tag{19}$$

where $h_c$ is equal to the ratio of the heat conductivity of the substrate material (AISI 1018: 51.9 W/(m K)) [40] and the height of the substrate (15 mm). The temperature data at the last layer and the current building layer measured by the IR camera is assumed to be known and is used for calculating $\mathcal{L}_d$ when training the PINN.

**Fig. 12** The workflow of processing the IR data

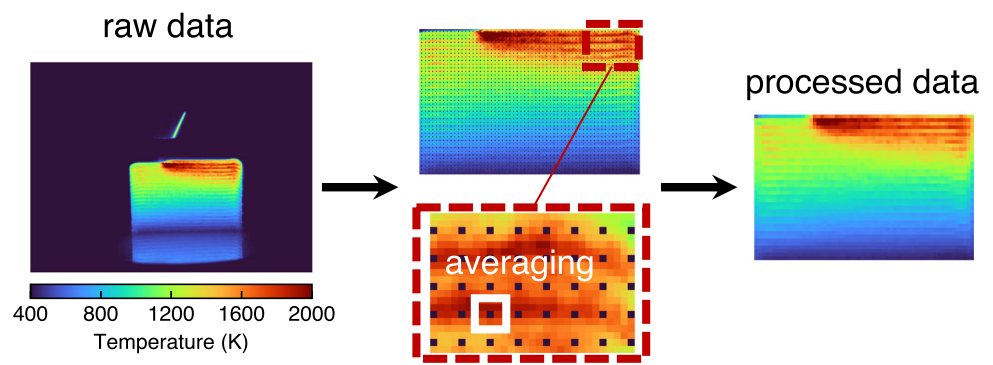

**Table 3** Parameters used in the PINN model for the thin wall example

| Parameter | Value | Unit |
|---|---|---|
| Density, $\rho$ | 8.19 | g/cm$^3$ |
| Heat capacity, $C_p$ | 0.00020465T + 0.38091 | J/(g K) |
| Heat conductivity, $k$ | 0.016702T + 5.5228 | W/(m K) |
| Convection coefficient, $h$ | 20 | W/(m$^2$·K) |
| Emissivity, $\varepsilon$ | 0.2 | – |
| Ambient temperature, $T_0$ | 298 | K |

A PINN model consisting of 3 hidden layers with 64 neurons in each layer was trained for 100,000 epochs using an Adam optimizer with a learning rate of 2e−4 to solve the full-field thermal history during the deposition of the last layer from the boundary conditions described above and the measured temperature data. The parameters used in the model are given in Table 3. Temperature-dependent heat capacity and thermal conductivity values of IN718 [41] were used.

Figure 14 depicts the results of the predicted full-field temperature from the trained PINN model compared to the measured IR data. The RMSE error between the predicted temperature field and the measured data from $t = 0$−7 s is 47.28 K. It can be observed that the PINN model can predict the full field temperature with good accuracy. This example demonstrates the effectiveness of the developed hybrid physics-based data-driven framework to arbitrarily fuse the experimental data into the physics-informed model for analyzing the thermal behavior in AM processes.

## 6 Conclusions and future work

In this study, a hybrid physics-based data-driven thermal model of metal AM processes based on PINNs is developed for predicting a full-field temperature history and identifying unknown material or process parameters from partially observed temperature data. Numerical and experimental examples are provided. The fundamental findings

and their corresponding impacts from the examples can be enumerated as:

(1) PINNs can be used to solve the forward problems in AM for predicting thermal history without labelled data. In the tested numerical case, the RMSE error compared to FEM simulation results was 14.07 K. This shows that PINN can be used as an effective surrogate model for the thermal behavior in AM processes.

(2) Adding auxiliary data can help the PINN model to solve forward problems with better accuracy and convergence speed. With the auxiliary data, the PINN model reached the same accuracy with only 1/3 of the number of epochs compared to that without auxiliary data. In the tested numerical case both PINNs trained with the clean data and noisy data achieved RMSE errors less than 4 K.

(3) Pretraining can largely accelerate the training of PINNs. In the numerical example, the PINN model started from a pretrained model took less than 1/5 of the number of epochs in the case of a randomly initialized model to reach the same accuracy. Thus, the developed PINN-based thermal model is suitable in the case when multiple simulations with different materials or process parameters are desired.

(4) The PINN-based hybrid thermal model can be used to identify unknown values of material and process parameters from partially observed temperature data. In the numerical example, the identified laser absorptivity, heat capacity and thermal conductivity were of less than 5% errors with the synthetic noisy IR images.

(5) In the experimental example, the full-field thermal history was successfully predicted from partially observed temperature data measured by the IR camera using the developed hybrid framework. The RMSE error between the prediction and measured data was 47.28 K. It is demonstrated that the hybrid framework enables arbitrary fusion of experimental data into the PINN model, which provides a flexible way to model the thermal behavior in AM.

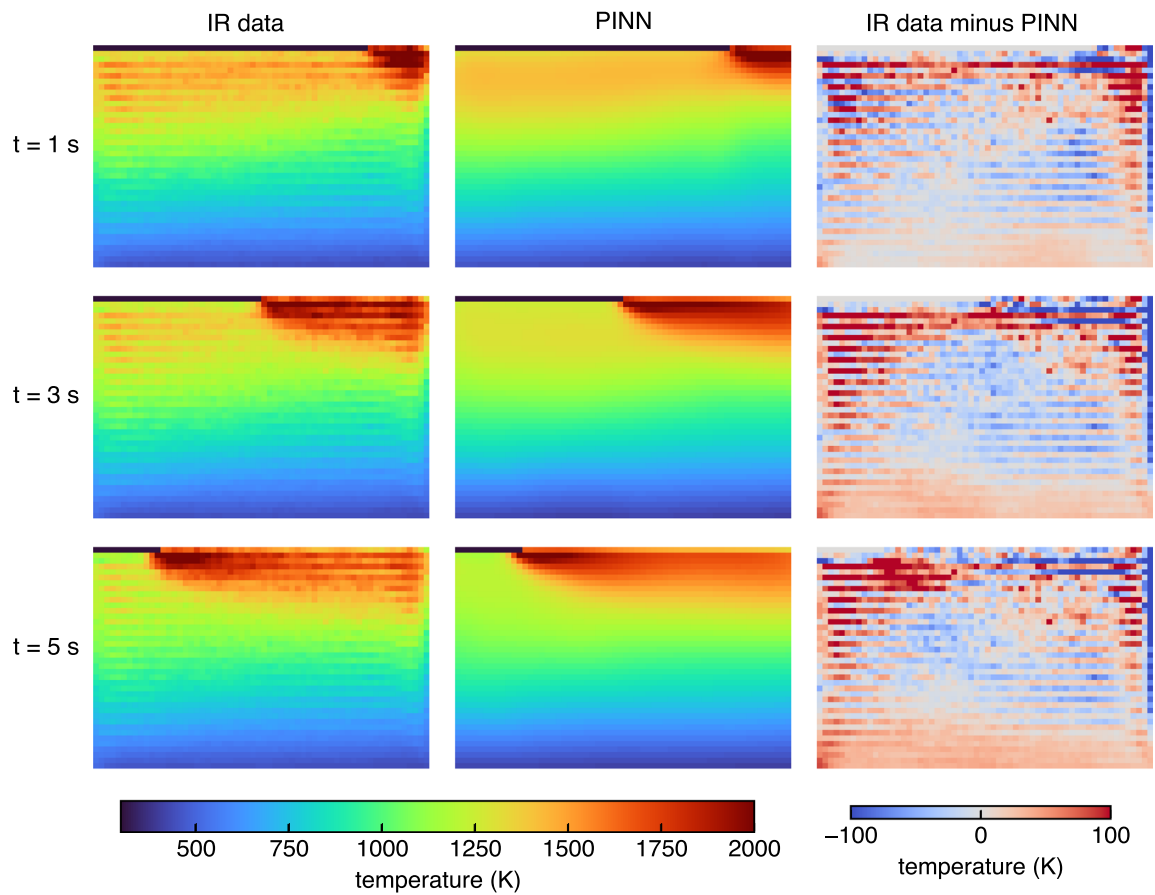

**Fig. 14** Results of full field temperature prediction: comparison of the measured data and prediction results

The current work, to the best of our knowledge, for the first time combines PINNs with experimental data for full-field temperature prediction and parameter identification in AM. Compared to conventional numerical methods, one limitation of the PINN model is that currently there is a lack of theory to guide the optimization of the NN architecture. Also, the requirement of training the PINN makes it much more computationally expensive than commonly used numerical methods when only a single simulation case is performed. However, because of the effect of pretraining and the ability to fuse experimental data with physics laws, the PINN model can be used in an offline-online fashion in AM applications to achieve efficient and accurate online prediction and control. In future work, more physics laws, e.g., the fluid flow and phase transitions, can be added to the developed hybrid framework. Since the current experimental measurements using the IR camera is not very accurate, more reliable measurement setups, e.g., the co-axial high-resolution Planck thermometry measurements [42], can be integrated with the PINN model. In additional to modeling the thermal behavior, the application of PINNs to model the microstructure evolution [43] and residual stress will also be developed. Moreover, since PINNs have been demonstrated to solve inverse design problems [44] successfully, we will also explore the possibility of using PINNs to design the process parameters for AM applications.

**Acknowledgements** This work was supported by National Institute of Standards and Technology (NIST)-Center for Hierarchical Material Design (CHiMaD) under grant No. 70 NANB19H005, and the Department of Defense Vannevar Bush Faculty Fellowship, USA N00014-19-1-2642.

**Data Availability** Our code and data can be found at https://github.com/ShuhengLiao/Physics_informed_AM.